\documentclass[conference]{IEEEtran}
\IEEEoverridecommandlockouts
\usepackage{cite}
\usepackage{amsmath,amssymb,amsfonts}
\usepackage{algorithmic}
\usepackage{graphicx}
\usepackage{textcomp}
\usepackage{xcolor}
\usepackage[numbers]{natbib}
\usepackage{booktabs}
\usepackage{subcaption}
\usepackage{multirow}
\def\BibTeX{{\rm B\kern-.05em{\sc i\kern-.025em b}\kern-.08em
    T\kern-.1667em\lower.7ex\hbox{E}\kern-.125emX}}

\DeclareMathOperator*{\argmax}{arg\,max}

\begin{document}

\title{
TEAMMix: Taxonomy Enrichment Augmentation and  Minority-augmented Mixing Strategy for LLM-enhanced Weak-Supervised Hierarchical Text Classification
\thanks{*Corresponding Author}
}

\author{
    \IEEEauthorblockN{Jian Zhang$^{a}$, Zhuohao Yang$^{b}$, Songlin Lei$^{b}$, Bangli Liu$^{c}$, Ziwei Wang$^{c}$, Xufeng Weng$^{c}$,  \\ Gehan Amaratunga$^{b}$, Yu Lin$^{b}$, Hongwei Wang$^{abd*}$}
    \IEEEauthorblockA{$^a$ School of Computer Science and Technology, Zhejiang University, Hangzhou, China}
    \IEEEauthorblockA{$^b$ ZJU-UIUC Institute, Zhejiang University, Haining, China}
    \IEEEauthorblockA{$^c$ Shaoxing K3i Technology Co. Ltd}
    \IEEEauthorblockA{$^d$ State Key Laboratory of CAD\&CG, Zhejiang University, Hangzhou, China}
    \IEEEauthorblockA{\{jianzhang.22, hongweiwang\}@zju.edu.cn}
}

\maketitle

\begin{abstract}
Hierarchical Text Classification (HTC), as a critical text mining task, faces challenges such as complex label hierarchies and class imbalance. Existing methods based on large language models (LLMs) struggle to be efficiently applied to this task due to issues like lengthy prompts and loss of label structural information. To address these limitations, this paper proposes a weakly supervised HTC framework enhanced by LLM-based data augmentation. The framework first enriches the label hierarchy semantically through keyword generation and corpus mining, thereby enhancing the model's understanding of labels. Subsequently, it guides the LLM to generate pseudo-samples to mitigate the long-tail problem, and employs a Gaussian mixture model for confidence-based resampling to optimize the quality of generated data. Experimental results demonstrate that the proposed method effectively improves the reliability of LLM-generated pseudo-labels and significantly enhances classification performance on fine-grained and imbalanced datasets.
\end{abstract}


\section{Introduction}
Hierarchical text classification is a fundamental research area in Natural Language Processing (NLP) and web text mining, primarily focusing on categorizing web pages or documents containing multiple hierarchical label systems by assigning one or several labels to document content. Compared with traditional text classification tasks with single-layer label systems, vertical hierarchical text classification typically features larger label scales, more refined categorization systems, and imbalanced sample distribution across different label categories—known as the long-tail effect. This classification approach has numerous practical applications, including web content organization, semantic indexing of articles, and request categorization.

The core challenge in vertical hierarchical text classification lies in enabling models to comprehend large-scale label hierarchies, leverage semantic distinctions between labels, and align labels with textual content. Conventional approaches rely on fully-supervised or semi-supervised learning using extensive human-annotated data, which is often costly and difficult to scale. Consequently, unsupervised or weakly-supervised vertical classification methods have gained prominence, utilizing keyword information in texts and self-training pseudo-labeling techniques to achieve classification with minimal samples or using only the label hierarchy structure.

With the emergence of large language models (LLMs) like GPT and Gemini, significant progress has been made in flat text classification tasks within short timeframes. These LLM-based methods can achieve near state-of-the-art performance in text classification without requiring massive labeled datasets. However, in vertical hierarchical text classification, the enormous label quantities and complex hierarchical structures pose challenges: overlong prompt templates may lead to substantial loss of textual and structural information, while reducing the model's ability to distinguish characteristic features, ultimately hindering effective prompt construction and efficient classification in Hierarchical Text Classification (HTC). Current research applying large models to vertical text classification has adopted strategies like label hierarchy enrichment and label system representation generation to enhance weakly-supervised models' understanding of label characteristics and improve performance. Nevertheless, these methods still lack comprehensive verification and analysis regarding the reliability of pseudo-label data generated by large models.

\begin{figure*}
    \centering
    \includegraphics[width=0.85\textwidth]{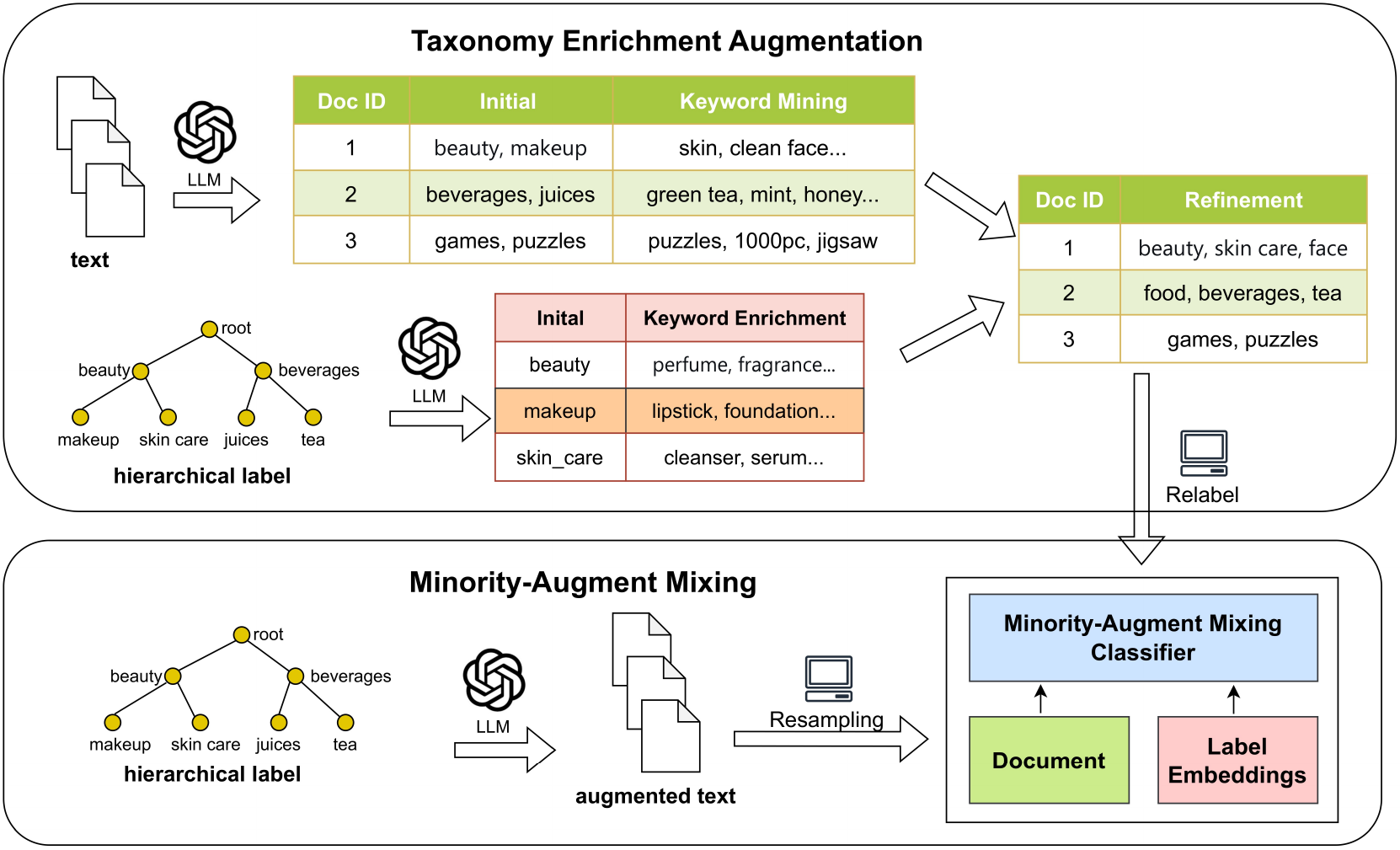}
    \caption{A overview of the TEAMMix framework}
    \label{fig:framework}
    \vspace{-2em}
\end{figure*}


In this study, we employ a LLM-enhanced strategy based on taxonomy enrichment and a minority-augmented mixing strategy to filter pseudo-samples and labels to improve the performance in weakly-supervised hierarchical text classification tasks. The framework is shown in Fig. \ref{fig:framework}. First, we utilize LLMs to directly generate semantic enrichment keywords based on the hierarchical label. We produce a set of keywords to enhance the model semantic representation of the label hierarchy. A tree-based search strategy is adopted to determine categories, combined with semantic analysis methods to extract indicative topic words from the corpus. The keywords generated by LLMs and those mined from the corpus are used to generate representations of category labels, establish initial core categories, and construct the label hierarchy structure. Second, guided by the obtained label hierarchy structure, we use LLMs to generate the corresponding pseudo-samples. This enhances the proportion of minority-class samples by utilizing the augmented text data containing pseudo-samples. A Gaussian mixture model (GMM) is then applied to perform confidence resampling of generated pseudo samples, enhancing the semantic relevance between pseudo samples and labels.

The main contributions of this paper are as follows:
\begin{itemize}
    \item We propose an LLM-enhanced strategy that requires only category labels for hierarchical text classification.
    \item We introduce a minority-class augmented sampling strategy to screen and refine the LLM-generated samples.
    \item Experimental results validate the effectiveness of our method demonstrating improved quality of LLM-generated pseudo samples and labels, and alleviating the scarcity of fine-grained samples.
\end{itemize}

\section{Related Work}

\subsection{Weak-Supervised Classification}
Weakly-supervised text classification reduces reliance on manually annotated samples by training model weights with limited supervisory information. Current approaches primarily include: knowledge base-based distant supervision \cite{distantSupervised1, distantSupervised2, distantSupervised3}, keyword-induced pseudo labeling \cite{pseudolabel1, pseudolabel2, pseudolabel3}, and heuristic strategy-based methods \cite{heuristicStrategy1, heuristicStrategy2, heuristicStrategy3}. Recent studies have proposed minimal weak supervision techniques that utilize only label category information without any sample-label annotations, enabling classification model training through pseudo-labeling strategies.

\subsection{Hierarchical Text Classification}

Hierarchical text classification builds upon traditional flat text classification by leveraging the semantic relationships within label hierarchies to achieve structured categorization. Existing methods can be broadly divided into two categories, local classification approaches \cite{local1, local2, local3} and global classification approaches. The local approaches train multiple local classifiers to handle different levels and nodes of the hierarchy. The global approaches integrate hierarchical label structures into classifiers through techniques such as recursive regularization \cite{recursiveRegular} or a joint document label embeddings space \cite{jointDocumentLabel}. Recent studies \cite{LLMHTC1, LLMHTC2} exploring LLMs for hierarchical text classification reveal that LLMs still face significant limitations in comprehending complex hierarchical label structures. 

\subsection{Context Enrichment for LLMs}
Contextual Enrichment based on LLMs can address the inefficiency and limited scalability of contextual models in dynamic environments, enabling more effective capture of semantic relationships in data, including instruction-tuned error data identification \cite{LLMCleanContext} and semantic aware augmentation to explore modify and repair dataset iteratively \cite{LLM4dataexplore}.

\section{Method}

Weakly supervised vertical text classification tasks typically use a minimal set of label names as supervisory information. Through training, the model learns to categorize texts into the corresponding label system. The input for this task includes an unlabeled text corpus $D={d_1, d_2, ..., d_n}$ and a hierarchical label system $T = (N, E)$, where $c_i \to N$ represents the target categories in the label system, and each edge $<c_i, c_j> \to E$ denotes the hierarchical relationship between label categories. The objective is to train a multi-label text classifier f() that maps a text $d$ to a corresponding binary vector $f(d) = [h_1, h_2, ..., h_n]$, where $h_i=1$ indicates that d belongs to category $c_i$, and $h_i=0$ indicates it does not belong to that category.

\subsection{Taxonomy Enrichment Augmentation}

Through carefully designed prompts for large language models, a set of discriminative keywords is generated for each category to better distinguish between semantically similar classes. A top-down tree search algorithm is then utilized to match the most relevant candidate keywords for each text segment. By leveraging the category-keyword mapping, the initial core categories for the texts are established.

Subsequently, the semantic representation of labels is enhanced by mining keywords from the target corpus that demonstrate relevance to specific categories. Through comprehensive analysis of keyword popularity, discriminability, and semantic similarity within the corpus, the most representative terms for each category and their sibling categories are systematically filtered. 

Popularity is a frequency-based metric that measures how common a word is. The more frequently a word is used, the higher its popularity. It is mathematically represented as follows:
\begin{equation}
    \begin{aligned}
        pop(t,c) = log(1+df(t, D_c^0)),
        \label{eq:pop}
    \end{aligned}
\end{equation}

where $t$ is target word, $df(t, D_c^0)$ is the number of documents in $D_c^0$ that mention $t$.

Distinctiveness is a metric that evaluates how well a word represents and distinguishes a category. In this context, it is primarily used to select words that capture the core concepts represented by category labels. The calculation method is as follows:
\begin{equation}
    dist(t,c,c_p) = \frac{exp(BM25(t, D_c^0))}{1+\Sigma_{c' \in sib(c,c_p)}  exp(BM25(t, D_c^0))}
    \label{eq:dist}
\end{equation}
where $sib(c, c_p) = \{ c' \in N | <c_p, c'> \in E \}$ is a class $c \in N$ and its siblings to one of its parents $c_p$. $BM25(t, D_c^0)$ denotes the $BM25$ relevant function in $t$ and $D_c^0$.

Semantic Similarity is a commonly used metric in text matching. We compute the cosine distance between word embeddings generated by BERT to evaluate similarity, represented by the following formula:
\begin{equation}
    sim(c,d) = \max \limits_{t \in T_c^{LLM}} cos(\overrightarrow{t}, \overrightarrow{d})
    \label{eq:sim}
\end{equation}
where $T_c^{LLM}$ denotes prompting an LLM to enrich label taxonomy with a set of key terms for class $c$, $\overrightarrow{t}$ and $\overrightarrow{d}$ denote vectors representation by a pretrained semantic encoders, e.g. Sentence-BERT \cite{SentenceBERT}.

This dual enhancement strategy - combining LLM-generated keywords with corpus-mined terms - enriches the semantic representation of keywords for each category, thereby significantly improving the recognition capability for fine-grained or long-tail categories.

\begin{equation}
    T_c = ( \bigcup_{c_p \in Par(c) } {T(c,c_p))} T_c^{LLM}
\end{equation}
where $T_(c, c_p)$ denotes corpus-extracted terms with highest affinity scores between class $c$ and its parents $c_p$, $T_c$ denotes the union with corpus-based terms and LLM-generated terms.

After obtaining the keyword-enhanced representations for all categories, the cosine similarity between pre-trained sentence encoders and category representation vectors is employed to further evaluate matching accuracy. By identifying the maximum divergence point in similarity rankings, high-confidence text-category pairs are selected as pseudo-labeled data for subsequent training phases, substantially enhancing the quality of pseudo-labels for textual data.
\begin{equation}
\begin{aligned}
    \mathrm{conf}_i &= \mathrm{diff}^i(m_i) \\
    \mathcal{C} &= {c_1^i, c_2^i, ... c_{m_i}^i} \\
    m_i = \argmax \limits_{j \in \{ 1, ..., |C|-1 \}} &\cos(\overrightarrow{d_i}, \overrightarrow{c_j^i}) - \cos(\overrightarrow{d_i}, \overrightarrow{c_{j+1}^i}) \\
\end{aligned}
\end{equation}
where $m_i$ is the position of highest similarity difference with its next in this list.

\subsection{Minority-augmented Mixing}

In addressing the issue of label scarcity, we introduced a strategy combining LLM-generated data with hybrid sampling. By leveraging the taxonomy tree, we generated multiple pseudo-texts for each path to ensure every label has corresponding textual data. During the classifier training phase, all generated texts undergo dynamic confidence-based sampling for screening and refinement. The mathematical representation of this dynamic sampling based on prediction confidence is as follows:
\begin{equation}
    \begin{aligned}
        p_{sampling}((d,c); \mathcal{D}) = \frac{1/Score(d,c)}{\Sigma_{d',c'}\to \mathcal{D}^{1/Score(d', c')}}
    \end{aligned}
\end{equation}

The confidence score can be calculated using the following method:
\begin{equation}
    \begin{aligned}
        Score(d,c) = \sum_{k=1}^{K} 1_{[c_k=1]} P(k) + 1_{[c_k=0]} A(k)
    \end{aligned}
\end{equation}
where $P(k)$ and $A(k)$ denote presence and absence of the $k$-th class, $1_{[c_k=1]}$ and $1_{[c_k=0]}$ are the one-hot score with confidence score for each positive and negative instance. 

We employ GMM for loss modeling to achieve label screening as shown below:
\begin{equation}
    \begin{aligned}
        p_{\mathcal{G}}(d, c_k=l) = \frac{\mathcal{G}_{(c_k=l)} (BCE(f_{(d,c_k=l)|g}) \mathcal{G}_{c_k=l}(g)}{\mathcal{G}_{c_k=l} (BCE(f_{(d,c_k=l}))}
    \end{aligned}
    \label{eq:resampling}
\end{equation}
where $\mathcal{G}$ is GMM, $BCE$ is binary cross entropy function, $f_{(d,c)}$ is the prediction confidence with document and label, $g$ denotes a modality for small loss label. Thus, a label with $p_\mathcal{G} > 0.5$ would be marked as clean label.

We modify the given label which is not selected as a clean label but model exhibits high confidence in predictions. we ensemble the prediction confidences on two augmented instances from the original instance $d$ whose $p_\mathcal{G} \leq 0.5$, the $k$-th label is relabeled as follows:
\begin{equation}
    \begin{aligned}
        1/2 f_{(aug_1(d), c_k)} + f_{(aug_2(d), c_k)} &> \epsilon & \Rightarrow  y_k = 1 \\
        1/2 f_{(aug_1(d), c_k)} + f_{(aug_2(d), c_k)} &< 1 - \epsilon & \Rightarrow y_k = 0 \\
    \end{aligned}
    \label{eq:relabeling}
\end{equation}
where $\epsilon$ is the confidence threshold for relabeling.

The remaining labels are treated as ambiguous labels and are reweighted in the loss calculation. The loss function can be expressed as:
\begin{equation}
    \begin{aligned}
        L = \frac{1}{B_{mix}} \sum_{(d^{mix}, c^{mix}) \in B_{mix}} & (\sum_{k \in C} BCE(f_{(d^{mix}, c^{mix})}) \\ 
        + & \sum_{k \in U} BCE(f_{d^{mix}, c^{mix}}))
        \vspace{-0.5em}
        \label{eq:loss_function}
    \end{aligned}
\end{equation}
where $d^{mix}$ and $c^{mix}$ are the mixture sample from clean and unclean label, $C$ is a set of clean sample and $U$ is a set of unclean sample.

\section{Experiment and Analysis}

\subsection{Experiment Setup}

\subsubsection{Datasets}

\begin{table}[]
    \centering
    \caption{Dataset Statistic}
    \begin{tabular}{c|ccc}
    \toprule
    \textbf{Dataset} & \textbf{unlabeled train} & \textbf{test} & \textbf{labels} \\
    \midrule
    \textbf{Amazon-531 }& 29487 & 19865 & 531 \\
    \textbf{DBPedia-298} & 196665 & 49167 & 298 \\
    \bottomrule
    \end{tabular}
    
    \vspace{-1em}
    \label{tab:dataset}
\end{table}

\begin{table*}[]
    \centering
    \caption{Main result on Amazon-531 and DBPedia-296 datasets, the best score among zero-shot and weakly-supervised methods is bold. "*" indicates the results are directly from previous paper \cite{teleclass}. "-" represents MRR cannot be calculated.}
    \begin{tabular}{cl|cccc|cccc}
       \toprule 
       \multirow{2}{*}{\textbf{Supervised type}} & \multirow{2}{*}{\textbf{Methods}} & \multicolumn{4}{c|}{ \textbf{Amazon-531}} & \multicolumn{4}{c}{\textbf{DBPedia-296}} \\
       & & Example-F1 & P@1 & P@3 & MRR & Example-F1 & P@1 & P@3 & MRR \\
       \midrule
        \multirow{2}{*}{Zero-Shot} & Hier-0-shot-TC* & 0.4742 & 0.7144 & 0.4610 & - & 0.6765 & 0.7871 & 0.6765 & - \\
         & ChatGPT* & 0.5164 & 0.6807 & 0.4752 & -  & 0.4816 & 0.5328 & 0.4547 & -  \\
        \midrule
         \multirow{6}{*}{weakly-supervised} & Hier-doc2vec* & 0.3157 & 0.5805 & 0.3115 & - & 0.1443 & 0.2635 & 0.1443 & - \\
         & WeSHClass* & 0.2458 & 0.5773 & 0.2517 & - & 0.3047 & 0.5359 & 0.3048 & -  \\
         & TaxoClass-NoST* & 0.5431 & 0.7918 & 0.5414 & 0.5911 & 0.7712 & 0.8621 & 0.7712 & 0.8221  \\
         & TaxoClass* & 0.5934 & \textbf{0.8120} & 0.5894 & 0.6332 & 0.8156 & 0.8942 & 0.8156 & 0.8762   \\
         &  TELEClass & 0.6206 & 0.7875 & 0.6148 & 0.6623 & 0.8536 & 0.9007 & 0.8536 & 0.8726 \\
         & TEAMMix(Ours) & \textbf{0.6331} & 0.7739 & \textbf{0.6276} & \textbf{0.6656} & \textbf{0.8590} &\textbf{0.9260} & \textbf{0.8590} & \textbf{0.8796}  \\
        \midrule
        \multicolumn{2}{c|}{Fully-Supervised*} & 0.8843 & 0.9524 & 0.8758 & 0.9085 & 0.9786 & 0.9945 & 0.9786 & 0.9826 \\
        \bottomrule
    \end{tabular}
    
    \vspace{-1em}
    \label{tab:main_result}
\end{table*}


For our proposed method, we selected two public datasets from different domains for experiments. The statistical information of these datasets is presented in Table \ref{tab:dataset}.

\paragraph{Amazon-531 \cite{amazon531}}  Based on product reviews from the Amazon e-commerce system, with a three-level product category labeling system.

\paragraph{DBPedia-298 \cite{dbpedia296}} Comprises Wikipedia documents annotated with a three-level category labeling system.

\subsubsection{Baseline}

We have selected zero-shot learning-based and weakly supervised learning-based methods as baseline models to compare with our approach. Additionally, a classifier trained on complete data was used for control experiments.  
The baseline methods we selected are as follows: Hier-0-shot-TC \cite{hier-0shot-TC} and Hier-doc2vec \cite{hier-doc2vec} are the zero-shot methods, another type is weakly supervised method, e.g. WeSHClass \cite{weshclass}, TaxoClass \cite{taxoclass} and TELEClass \cite{teleclass}.






\subsubsection{Metric}

In this study, we adopt three evaluation metrics to assess model performance: Example-F1, Precision at k (P@k) (k=1,3) and Mean Reciprocal Rank (MRR). The definitions of these metrics are as follows:

\paragraph{Example-F1}

\begin{equation}
    \text{Example-F1} = \frac{1}{|\mathcal{D}|} \sum_{d_i \in \mathbb{D}} \frac{2|\mathbb{C}^{true}_{i} \cap \mathbb{C}_{i}^{pred}|}{|\mathbb{C}_{i}^{true}|+|\mathbb{C}_{i}^{pred}|}
\end{equation}


\paragraph{Precision at k} 
P@k is a ranking-based metric which evaluates the precision of top-$k$ classes by predicted.

\begin{equation}
    P@k = \frac{1}{k} \sum_{d_i \in \mathcal{D}} \frac{|\mathbb{C}_{i}^{true} \cap \mathbb{C}_{i,k}^{pred}|}{min(k, |\mathcal{}|)}
\end{equation}

\paragraph{Mean Reciprocal Rank}
MRR is a ranking-based metric which evaluates multi-label method performances with the inverse of true labels ranks within predicted classes.

\begin{equation}
    MRR  = \frac{1}{|\mathcal{D}|} \sum_{d_i \in \mathcal{D}} \frac{1}{|\mathbb{C}_i^{true}|} \sum_{c_j \in \mathbb{C}_i^{true}} \frac{1}{min\{k|c_j in \mathbb{C}_{i,k}^{pred}\}}
\end{equation}

\subsection{Implementation Details}
In the TEA module, our experimental setup follows \citet{teleclass} approach. We used sentence transformer as text encoder to construct the category-keyword mapping. ChatGLM-4-9B served as the base LLM model for taxonomy enrichment. BERT-base-uncased provided feature embeddings for calculating popularity, distinctiveness and similarity as Eq. (\ref{eq:pop}), (\ref{eq:dist}) and (\ref{eq:sim}). For the category-keyword mapping, we selected 20 different words as top-k keywords to enrich semantic representation for each category. In the MMix module, 5 distinct pseudo-samples per category were generated to address minority class sample scarcity. Confidence filtering used a beta distribution with $\alpha=3.0$ and a confidence threshold $\epsilon=0.95$. The final multi-label classifier configuration used BERT-base-uncased as the embedding layer, optimizer selected AdamW with $5e-5$ learning rate and batch size is 32. The experiments was conducted with 10 epochs. All experiments were conducted on a single GPU NVIDIA RTX 4090 24G.

\subsection{Main Results}

Table \ref{tab:main_result} presents the results of all baseline models and our proposed method. Overall, our method significantly outperforms other baseline models in unsupervised and weakly supervised scenarios across $Example-F1$, $P@3$, and $MRR$ metrics, demonstrating its effectiveness for hierarchical text classification tasks. While achieving competitive performance with TELEClass on $P@1$, our method surpasses it substantially on all other metrics. Notably, compared to TELEClass, our improvements primarily stem from the minority-class sample mixing enhancement strategy—though we only optimized the prompt content for pseudo-sample generation, the key performance gain originates from leveraging LLM-generated high-quality pseudo-samples through this enhanced mixing mechanism.



\subsection{Sensitive Analysis}
To evaluate the minority augmented mixing strategy method using GMM comprehensively, a corresponding hyperparameter sensitivity experiment was designed, focusing on two core parameters, the distribution control hyperparameter $\alpha$ and the confidence threshold $\epsilon$. While keeping other hyperparameter fixed, we individually adjusted each target parameter and validated their impact on model performance using the example-F1 metric on the Amazon-531 dataset.

\begin{figure*}[htbp]
    \centering
    \begin{subfigure}[b]{0.48\linewidth}
        \centering
        \includegraphics[width=0.95\linewidth, trim=0 1.5em 0 1.5em, clip]{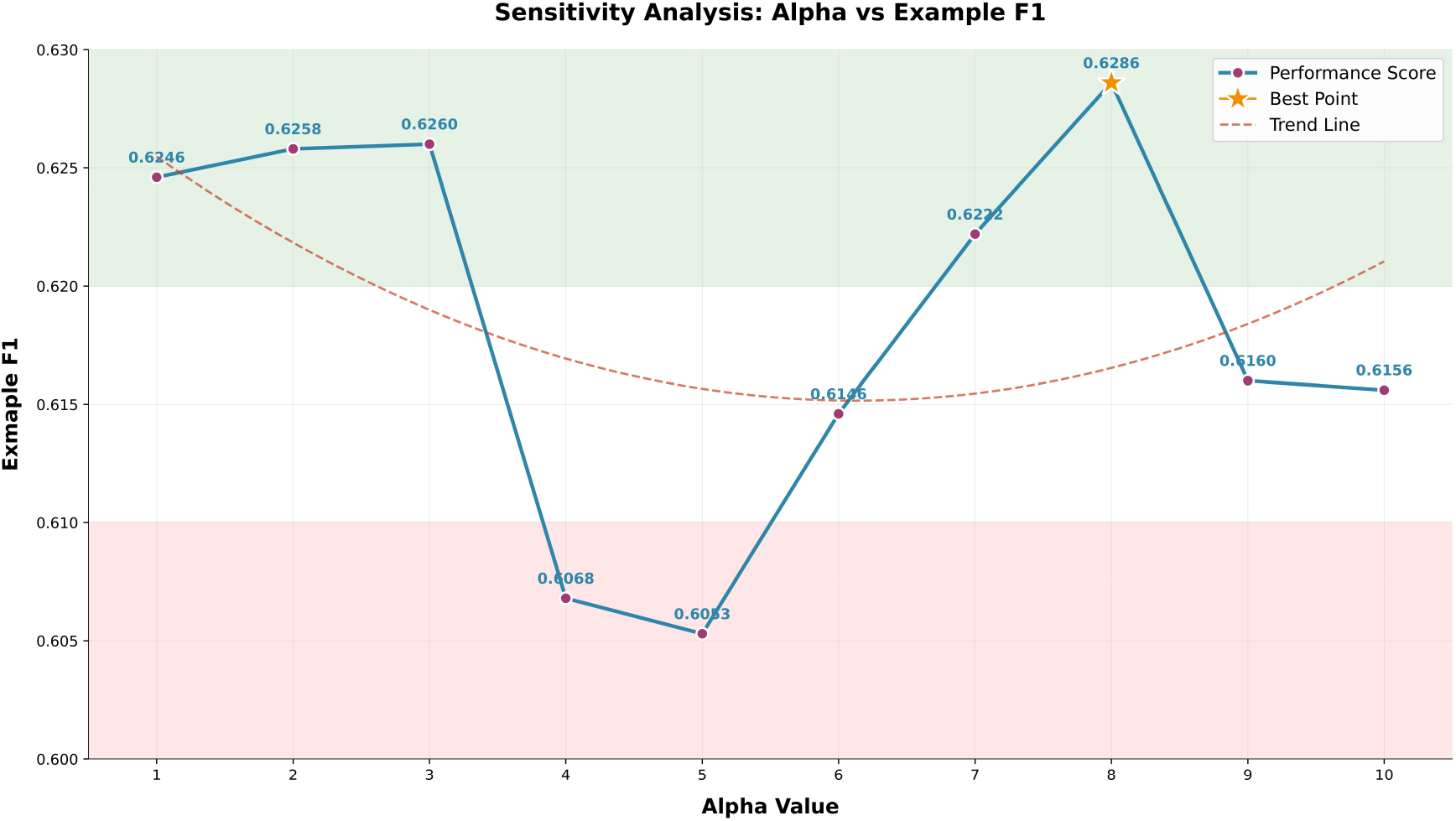}
        \caption{The sensitive curve for alpha}
        \label{fig:sensitive_alpha}
    \end{subfigure}
    \hfill
    \begin{subfigure}[b]{0.48\linewidth}
        \centering
        \includegraphics[width=0.95\linewidth, trim=0 1.5em 0 1.5em, clip]{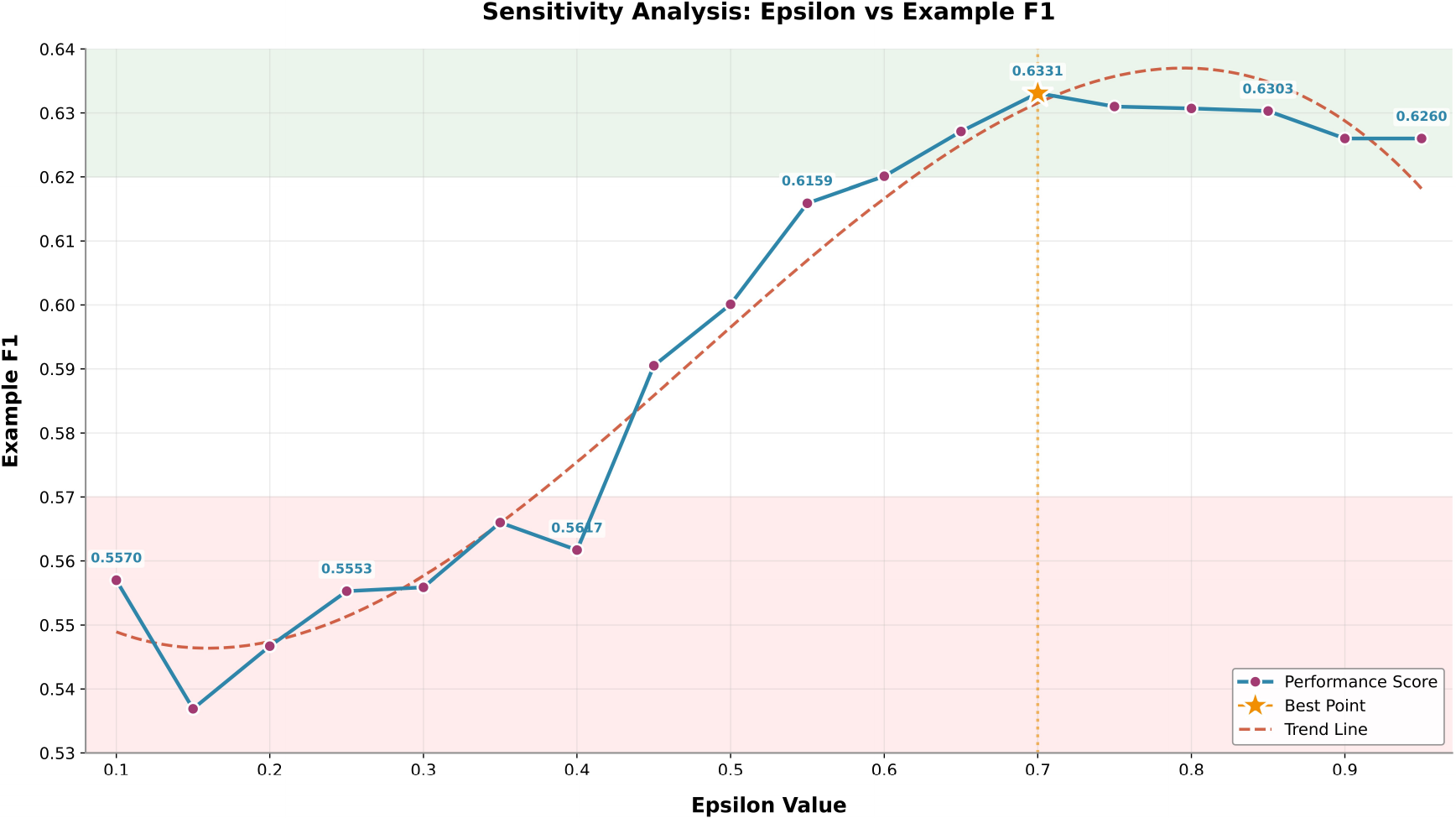}
        \caption{The sensitive cure for epsilon}
        \label{fig:sensitive_epsilon}
    \end{subfigure}
    \caption{The sensitive analysis}
    \label{fig:sensitive}
\end{figure*}


As shown in Eq. (\ref{eq:resampling}), the hyperparameter $\alpha$ influences the resamples mixing ratio for minority classes. The setting of this sampler parameter significantly affects both the diversity of random resampled examples and the contextual semantics of minority class samples. By adjusting $\alpha$, we validated its impact on actual model performance, with specific experimental results as Figure \ref{fig:sensitive}.

The confidence threshold $\epsilon$ serves as a hyperparameter to evaluate the filter whether the current model can accurately convey pseudo sample and label relationship. Through dynamic analysis of this parameter, we aimed to uncover valid information in pseudo samples generated by LLMs that might obscured by noise, thereby verifying the consistency between pseudo texts and their labels.

\subsection{LLM Prompt Comparison}

In this section of experiments, we further compared the method of direct classification using LLM prompts, evaluating performance with F1 and Precision@k metrics while also comparing the computational cost and time consumption across different models.

We used GPT-3.5 and GPT-4 as baseline comparison methods: For GPT-3.5, we provided prompts containing all categories and instructed the model to return the three most relevant labels. For GPT-4, we employed the same full-category prompt strategy but evaluated the model's Example-F1 and P@k metrics only on a 1000-samples test subset. We then extrapolated the required cost and time consumption for the complete dataset based on this subset.

As shown in Table \ref{tab:llm_prompt}, on the Amazon-531 dataset, our proposed TEAMMix method shows a slight performance decrease across all three metrics compared to GPT-4. However, when considering time and cost efficiency, our approach significantly outperforms GPT-4 direct prompt-based classification while maintaining comparable performance. This advantage becomes more pronounced as the dataset scales, highlighting the scalability challenge of prompt-based methods. These results demonstrate that our TEAMMix method is particularly suitable for applications with high real-time requirements.

\begin{table}[]
    \centering
    \caption{Result on zero-shot LLM prompting. "*" indicates the results are directly from previous paper \cite{teleclass}, "\textdagger" represents only use a subset from test about 1000 samples. }
    \vspace{-0.5em}
    \begin{tabular}{c|ccccc}
    \toprule
    \multirow{2}{*}{Methods} & \multicolumn{5}{c}{ Amazon-531} \\
     & Example-F1 & P@1 & P@3  & Cost & Time \\
    \midrule
    GPT-3.5*  & 0.5164 & 0.6807 & 0.4752 & \$60 & 240 mins \\
    GPT-4* \textdagger  & 0.6994 & 0.8220 & 0.6890 & \$800 & 400 mins \\
    TEAMMix &  0.6331 & 0.7739 & 0.6276 & $<$\$1 & 2 mins \\
    \bottomrule
    \end{tabular}
    \vspace{-1em}
    \label{tab:llm_prompt}
\end{table}


\section{Conclusion}
In this study, we propose a LLM-based taxonomy-enrichment and a minority-augmented mixing strategy to address hierarchical text classification tasks in weakly-supervised scenarios. By employing LLM-generated keywords and corpus-based term extraction methods, we enhance the semantic representation of the label taxonomy, thereby strengthening the model's understanding of label semantics. Additionally, we utilize a confidence-based approach combined with a minority-class-augmented mixing strategy to filter generated texts and labels, enabling more efficient and accurate acquisition of minority-class samples and label representation features. Experimental results validate the effectiveness of our proposed method, while ablation studies analyze the contribution of each module. Sensitivity analysis further verifies the operational range of the model's hyperparameters.

For future works, we plan to further explore LLM-enhanced low-resource text mining methods for hierarchical label taxonomies, such as nested named entity recognition. Additionally, the noise resistance of the generated texts and label keywords in the current model requires further improvement. Finally, this research could be extended to other annotation-scarce tasks, leveraging LLM-enhanced strategies to address challenges in low-resource environments with limited training data or more complex label taxonomy tasks.

\section*{Acknowledgment}
We would like to thank the anonymous reviewers for their valuable comments. This work is supported by National Key Research and Development Program of China (2024YFF0907803).

\bibliographystyle{IEEEtranN}  
\bibliography{ref}

\end{document}